\documentclass[hidelinks]{article}

\usepackage[preprint]{neurips_2019}

\usepackage[utf8]{inputenc} % allow utf-8 input
\usepackage[T1]{fontenc}    % use 8-bit T1 fonts
\usepackage{hyperref}       % hyperlinks
\usepackage{url}            % simple URL typesetting
\usepackage{booktabs}       % professional-quality tables
\usepackage{amsfonts}       % blackboard math symbols
\usepackage{nicefrac}       % compact symbols for 1/2, etc.
\usepackage{microtype}
\usepackage[ruled,linesnumbered]{algorithm2e}
\usepackage{graphicx}
\usepackage{float}

\usepackage[acronym]{glossaries}
\newacronym{rl}{RL}{Reinforcement Learning}
\newacronym{hrl}{hRL}{Hierarchical Reinforcement Learning}
\newacronym{ham}{HAM}{Hierarchies of Abstract Machines}
\newacronym{mdp}{MDP}{Markov Decision Process}
\newacronym{smdp}{SMDP}{Semi-Markov Decision Process}
\newacronym{dqn}{DQN}{Deep Q-Network}
\newacronym{ddpg}{DDPG}{Deep Deterministic Policy Gradient}
\newacronym{ppo}{PPO}{Proximal Policy Optimization}
\newacronym{trpo}{TRPO}{Trust Region Policy Optimization}
\glsdisablehyper

\begin{document}

\title{Fast Task-Adaptation for Tasks Labeled Using Natural Language in Reinforcement Learning}
\author{Matthias Hutsebaut-Buysse \and Kevin Mets \and Steven Latr\'e \and \\Department of Computer Science, University of Antwerp - imec \\ \texttt{\{firstname.lastname\}@uantwerpen.be}}

\maketitle

\begin{abstract}
    Over its lifetime, a reinforcement learning agent is often tasked with different tasks. How to efficiently adapt a previously learned control policy from one task to another, remains an open research question. In this paper, we investigate how instructions formulated in natural language can enable faster and more effective task adaptation. This can serve as the basis for developing language instructed skills, which can be used in a lifelong learning setting. Our method is capable of assessing, given a set of developed base control policies, which policy will adapt best to a new unseen task.
\end{abstract}

\section{Introduction}
\gls{rl} solves sequential decision making problems by utilizing a trial-and-error approach guided by a reward signal. \gls{rl} has achieved tremendous successes, especially in beating humans in games \citep{silver2018alphazero, jaderberg2019pbt_ctf} and robotics \citep{levine2016e2e_visuomotor}. However, RL also suffers form various open problems, such as its sample inefficiency. This sample inefficiency is often caused by reward function-specification. A sparse and delayed reward signal makes it difficult for the agent to experience, and learn from, meaningful reward signals.

Designing tasks suitable to solve with \gls{rl} algorithms is often challenging \citep{ng1999reward_shaping}, and mostly involves designing a task-specific reward function. A recent line of research, surveyed by \citet{luketina2019suvery_rl_nlp}, has proposed methods that allow task descriptions to be specified using natural language. However, such methods \citep{chevalierboisvert2019babyai} have proven to still be very sample inefficient, requiring the usage of up to 50 GPUs during weeks in order to learn relatively simple tasks.

One promising approach includes \citet{jiang2019hal}, which proposed to tackle this sample inefficiency by decomposing the problem into an hierarchical structure, guided by the compositional nature of natural language.

Humans follow a similar strategy, and when confronted with a new problem, humans are generally capable of forming \textit{intuitive theories} about how to tackle the problem at hand. These intuitive theories often consist of sequences of high-level actions (e.g. \textit{first go to store x}, then \textit{stop for gas before driving home}). An interesting approach to make \gls{rl} more sample efficient, would be to combine high level human intuitive theories, expressed using natural language, and low-level automated trial and error learning.

An essential part of such a symbiosis is the ability of an agent to quickly adapt from one task to a similar task. A human does not need to learn each individual task from scratch, but has a set of base strategies from which a new strategy can be quickly formed.

Current algorithms capable of quickly adapting their control policies to solve related tasks, mostly rely on intensive training using a diverse set of tasks, often guided by a curriculum of increasingly more difficult and diverse tasks \citep{bengio2009curriculumlearning}.

In this paper, we take a different approach and examine if we can facilitate fast task adaptation by utilizing semantic meaning from task descriptions formulated in natural language. Our method is capable of, given a set of pre-trained control policies, and a new previously unseen task, making a decision about which previously developed control policy will adapt best, in order to solve a new previously unseen task, solely from its instruction.

In the following sections of this paper, we first briefly review key research related to ours (Section~\ref{sec:related_work}). Section~\ref{sec:babyai} contains a description of the environment, and tasks we use to demonstrate our method. In Section~\ref{sec:method} we describe the proposed method. Section~\ref{sec:experiments} demonstrates experimentally how well our method is capable of performing task-adaptation in a simple environment.

\section{Related work}
\label{sec:related_work}

Our proposed method can be situated on the intersection of transfer learning and natural language usage in \gls{rl}. In this section, we first briefly review how our method relates to key research in transfer learning in \gls{rl}, how natural language has been used in \gls{rl}, and what research has been conducted on this intersection.

\paragraph{Transfer learning in reinforcement learning}
Utilizing knowledge gained from learning one task to another task has been a widely studied field. The goal of this field is to make \gls{rl} more sample efficient \citep{konidaris2006transfer_rl, taylor2009transfer_rl_survey}. Common approaches include to train the agent on multiple tasks \citep{hessel2019popart}, or to construct parameterized policies \citep{schaul2015uvfa, andreas2016policy_sketches, oh2017zeroshottaskgeneralization}, which can be configured to perform new tasks. An alternative approach consists of learning inter-task mappings \citep{taylor2007representationtransferreinforcement}, based on task similarities. Our method similarly is capable of detecting task similarities, using additional information captured in task descriptions.

\paragraph{Language instructions in reinforcement learning}
Recent advances in \gls{rl}, surveyed by \citet{luketina2019suvery_rl_nlp}, have demonstrated the usage of natural language in order to build models capable of capturing domain knowledge. 

A commonly used approach consists of directly embedding both visual observation and language instruction in order to train a control policy \citep{hermann2017grounded, misra2017mapping, chevalierboisvert2019babyai}. Alternatively, \citet{goyal2019rewardshapelanguage} uses natural language reward shaping, by predicting if an action in a trajectory matches a task description. \citet{jiang2019hal} explores the compositional structure of natural language in order to train a hierarchical algorithm, capable of discovering abstractions that generalize over different sub-tasks using language instructions. However, current approaches commonly heavily depend on large amounts of human labeled data and hand-designed policies. In this context, our method can reduce the dependency on expensive human labeling by providing fast task-adaptation.

\paragraph{Transfer learning guided by language in reinforcement learning}

\citet{co-reyes2018metalearning} proposed a meta-learning algorithm capable of utilizing corrective instructions formulated in natural langue in order to facilitate task-adaptation.

Most similar to our research, is the work done by \citet{narasimhan2018language_transfer}, which includes a way to use entity descriptions in natural language as a layer of abstraction, in order to facilitate transfer of an \gls{rl}-policy, to a new environment.

\section{BabyAI environment}
\label{sec:babyai}

\paragraph{Environment}
In order to demonstrate the capabilities of our method, we make use of the \textit{BabyAI environment} proposed by \citet{chevalierboisvert2019babyai}. In this environment, the agent is tasked with completing various tasks in a 2D gridworld. The environment supports multiple rooms, but for our preliminary experiments, we only consider a single room, and use the \textit{goto} and \textit{pickup} tasks. The task the agent is charged with, is described using a synthetic \textit{baby language}. The pixels of the screen, together with this instruction, form the observation of the agent. The environment supports partial observability of the state. However for our experiments we use the fully observable configuration. The action-space we consider for our experiments consists of moving forward, turning left/right, object-pickup/drop, opening doors, and a \textit{finish} action. Notice that in order to solve the \textit{goto} and \textit{pickup} tasks, only a subset of the action-space is required.

\begin{figure}[ht]
    \centering       
    \includegraphics[scale=0.25]{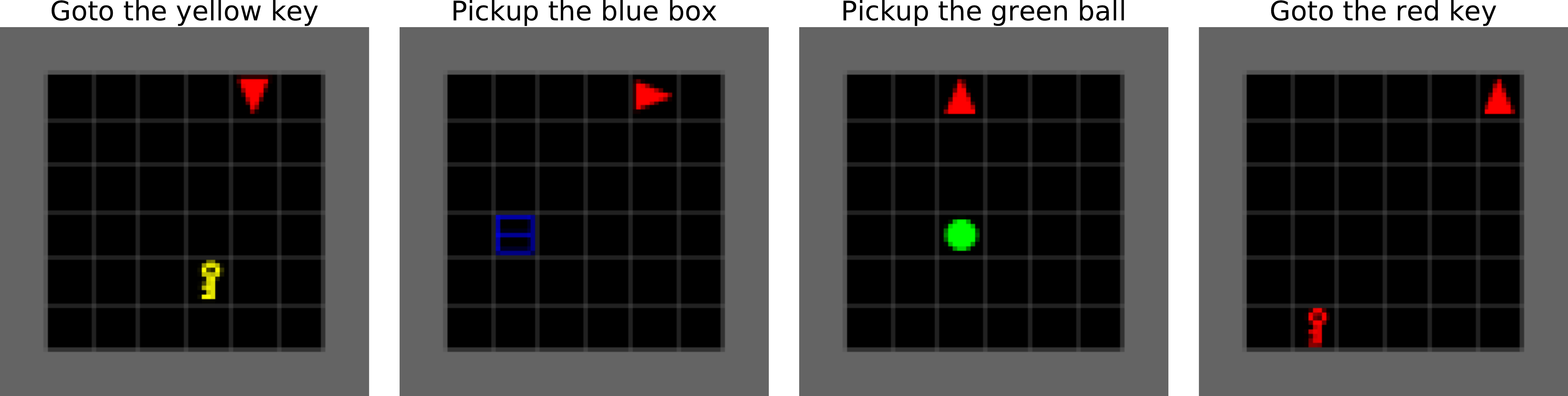}
    \caption{Sample tasks taken from the BabyAI environment \citep{chevalierboisvert2019babyai}. The current position of the agent is represented with a red triangle.}
    \label{fig:env}        
\end{figure}

In this environment, the reward-signal is only sparsely observed, as the agent only receives a reward upon task completion. A few example tasks are presented in Figure~\ref{fig:env}.

\paragraph{Vocabulary}
The instructions used in the BabyAI environment are all generated using the proposed \textit{Baby Language}. This language consists of a small vocabulary, but can be used combinatorially to express a relatively rich set of different tasks. 

Instructions we use in our transfer experiments follow the same \textit{verb}, \textit{object color}, \textit{object} pattern (e.g. \textit{pickup the yellow box}). The following words make up the vocabulary used in our experiments:

\begin{itemize}
    \item \textbf{Verbs}: pickup, goto
    \item \textbf{Objects}: box, key, ball
    \item \textbf{Colors}: blue, red, green, yellow
\end{itemize}

In total, this allows us to express 24 different tasks. While the BabyAI platform is a great platform to demonstrate the qualities of our method, our method is not environment-specific, and we plan to extend this research to multiple environments.

\section{Method}
\label{sec:method}

The main idea of our approach is to utilize a limited set of pre-trained base control policies. When confronted with a new task, described using natural language (the transfer instruction), the best base policy is selected and the new task is learned based on this base policy.

As such our method consists of two parts: the first part is a pre-training step, while the second part deals with the effective task-adaptation. A pseudo-code summary of our method can be found below in Algorithm~\ref{alg:summary}.

\begin{algorithm}[H]
    $\alpha$: $k$ instructions sampled from the set of possible instructions $Z$ \\
    $\beta$: $p$ instructions sampled from the set of possible instructions $Z$ \\

    \ForEach{instruction $z_i \in \alpha$} {
        Train base policy $\pi_i$ until convergence \\
        \ForEach{instruction $z_j \in \beta$} {
            Sample task-adaptation during $n$ training steps, from base policy $\pi_i$ (with instruction $z_i$) to task $z_j$
        }
    }
    Train the transfer model 

  \caption{Summary of our task-adaptation method}
  \label{alg:summary}
\end{algorithm}

\subsection{Pre-training base control policies}
In this pre-training phase, we first train a set of $k$ base control policies $\{\pi_{0},...,\pi_{k}\}$. A control policy $\pi_{i}(s_t)$ determines the action $a$ an agent takes, based on the state $s_t$ the agent resides in.

Each base control policy should reliably be able to perform one instruction $z_{i}$. This task instruction is expressed in natural language (e.g. \textit{go to the blue ball} or \textit{pickup the yellow key}). Training base control policies can be done using any \gls{rl} algorithm. In this preliminary research, the set of possible instructions $Z$ is limited. This is due to the fixed vocabulary described in Section~\ref{sec:babyai}. The amount of pre-trained control policies should be sufficiently large, but smaller than the entire set of possible instructions ($k \ll |Z|$).

For a base control policy to facilitate efficient task-adaptation, it is beneficial to make slight adaptations to the environment. An example of such variations includes spawning the agent in a different position after each iteration.

Our method can be used with a fixed number of base control policies, which are trained during a single pre-training phase. However, our method can also be extended to work in an iterative fashion. In this iterative approach, the agent starts with a small set of $k$ pre-trained base control policies. When confronted with a new task, our method is used to determine the best base control policy to facilitate task-adaptation (e.g. $\pi_i$). After training the new policy $\pi_j$ by adapting the selected base control policy $\pi_i$, the new policy $\pi_j$ can be added to the set of base control policies. This will allow executing more efficient task-adaptations, as more base control policies become available.

In the proposed method, we select $k$ instructions $\{z_{0},...,z_{k}\}$ to train base control policies $\{\pi_{0},...,\pi_{k}\}$, from a uniform random distribution. However, an interesting extension to this method might be to select base control policies based on a more advanced selection objective. For example, maximizing distance between the task instructions (in a language-embedding).

\subsection{Sampling task-adaptations}
The second phase of our method consists of utilizing the developed base control policies in order to sample a limited number of task adaptations. A single task adaptation sample $\langle \pi_i, z_j \rangle$ consists of taking a fully developed base control policy $\pi_i$, and using it to perform a new instruction $z_j$, different from the one it was trained on. An example of such a sample would include to start from a policy trained on an instruction \textit{go to the yellow box}, and ask it to perform a different task, such as \textit{pickup the yellow box}.

A task-adaptation from one policy to a new one is done by loading the parameters of the base policy as the initialization of the new policy we want to develop. Training can be performed using any \gls{rl}-algorithm. During this sampling phase the policy does not need to converge. Training only needs to happen for a limited number of $n$ steps. This amount of required steps is significantly lower than fully developing the policy. After the sampled task adaptation has been executed for $n$-steps, we measure the performance. This can be done by, for example, calculating the success rate of the agent satisfying the instruction over the last 100 iterations. Table~\ref{table:transfer_samples} contains a few examples of this sampling process.

\begin{table}[H]
    \centering
    \begin{tabular}{lll}
    \toprule
    \textbf{Base control policy instruction} & \textbf{Transfer instruction} &  \textbf{Measured performance} \\
    \midrule
    Pickup the red ball & Goto the green key & 0.91 \\
    Pickup the red ball & Goto the red ball & 0.76 \\
    Goto the yellow box &  Goto the green key & 0.86 \\
    Goto the yellow box &  Goto the red ball & 0.86 \\
    \bottomrule
    \\
    \end{tabular}
    
    \caption{Example task-adaptation sampling results ($k=2$ base policies, $p=2$ transfer instructions). The measured performance is calculated as the success rate of the 100 last episodes, after $n$ training steps. Displayed task performance is exemplary.}
    \label{table:transfer_samples}   
\end{table}

For each base control policy, we randomly select $p$ different tasks from $Z$ to sample task adaptation. So in summary, our method requires running $k \times p$ task adaptation samples, each consisting of $n$ training steps. Similarly to the selection of the base control policies, we leave a more advanced sampling strategy as future work.

This sampling method allows the generation of a dataset that can be used to generalize expected task adaptation over unseen tasks. Furthermore, the resulting policies, which were partially developed during the sampled task adaptations, could be used by the agent to further develop, when tasked with the linked instruction.

\subsection{Training the transfer-model}
In the next stage of our method, we train a binary classification model $f(z_{x},z_{i},z_{j}) \rightarrow \{1, 0\}$ in order to generalize the perceived task adaptation.

The input of the proposed model consists of a concatenation of the sampled transfer instruction $z_{x}$, combined with the instructions attached to two sampled base policies ($z_{i}$ and $z_{j}$). The output of the model consists of a single binary output. This output is trained to be positive, if the first base policy with instruction $z_{i}$ performed better during transfer sampling than the second base policy with instruction $z_{j}$. An example dataset is presented in Table~\ref{table:transfer_dataset}

\begin{table}[H]
    \centering
    \begin{tabular}{llll}
    \toprule
    \textbf{Instruction $z_{x}$} & \textbf{Transfer instruction $z_{i}$} & \textbf{Transfer instruction $z_{j}$} &  \textbf{Class} \\
    \midrule
    Goto the green key & Pickup the red ball & Goto the yellow box & 1 \\
    Goto the red ball & Pickup the red ball & Goto the yellow box & 0 \\
    \bottomrule
    \\
    \end{tabular}
    
    \caption{Example input dataset, used to train the transfer-model.}
    \label{table:transfer_dataset}   
\end{table}

\begin{figure}[H]
    \centering
        \includegraphics[scale=0.50]{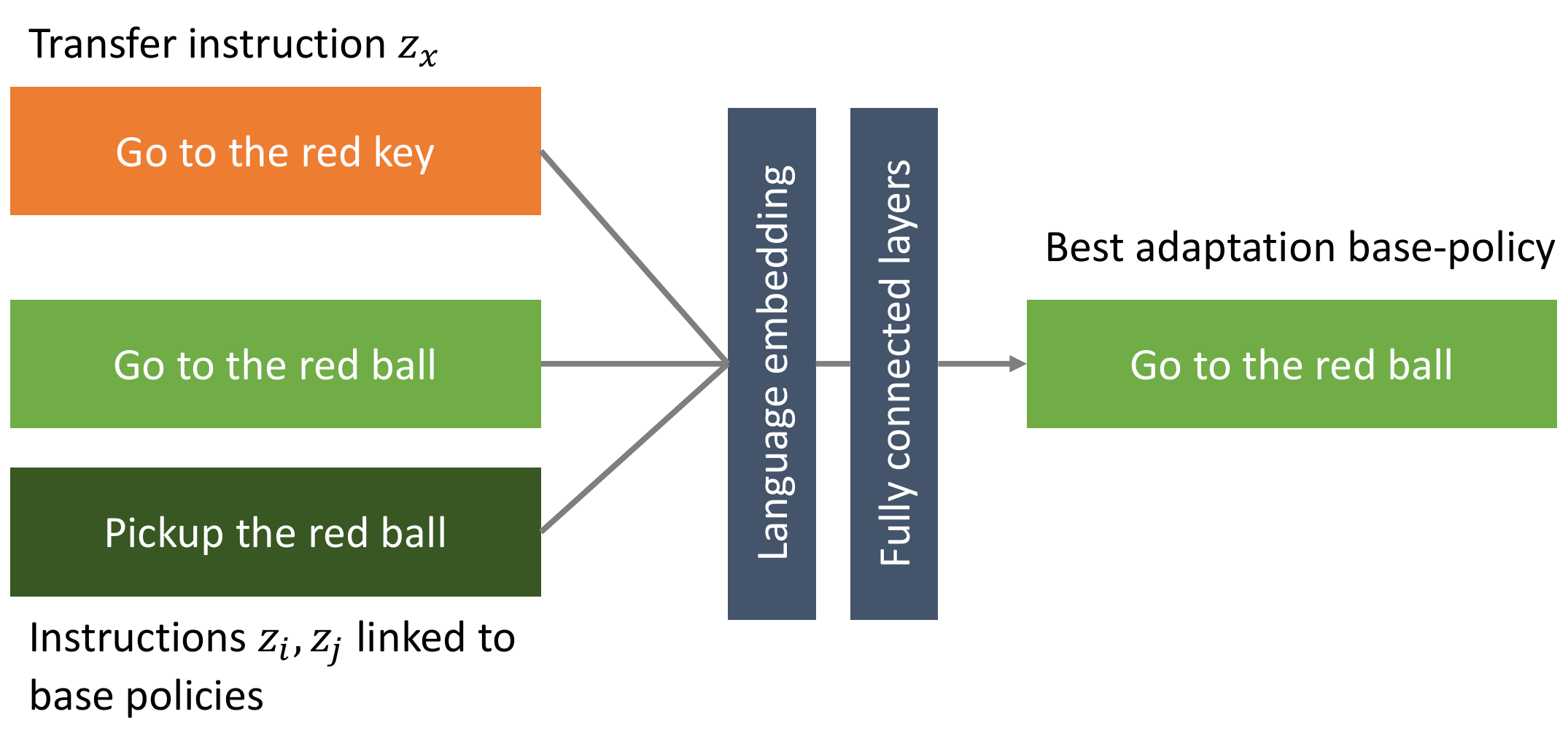}
        \caption{The high level transfer-model architecture. The input consists of a concatenation of the transfer instruction $z_{x}$, and instructions $\langle z_{i}, z_{j}\rangle$ linked to two base policies. A language-embedding layer is used in order to learn a task specific language-embedding. This layer is followed by a set of fully connected layers which finally output a binary variable.}
        \label{fig:model}
    \end{figure}

In order to work directly with instructions in natural language, a language embedding is used. This embedding is trained end-to-end, and thus is specifically trained to encode instructions based on their transfer capabilities.

\subsection{Transfer-model usage}
The resulting transfer-model can be used when the agent is confronted with a new task, it currently has no developed base control policy for. Given a set of labeled base policies, and a task instruction, the various possibilities can be tested in order to make an assessment about which base policy will result in the fastest task-adaptation. 

\section{Experiments}
\label{sec:experiments}

    \subsection{Task-adaptation in the BabyAI environment}
    In order to find out whether patterns can be discovered in task adaptations using instructions expressed using natural language, we performed a large set of transfer experiments in the BabyAI environment. In this experiments, we wanted to find out which parts of the instructions (verb, object, color) matter in making efficient task adaptation decisions.

    \begin{figure}[p]
        \centering
        \includegraphics[scale=0.45]{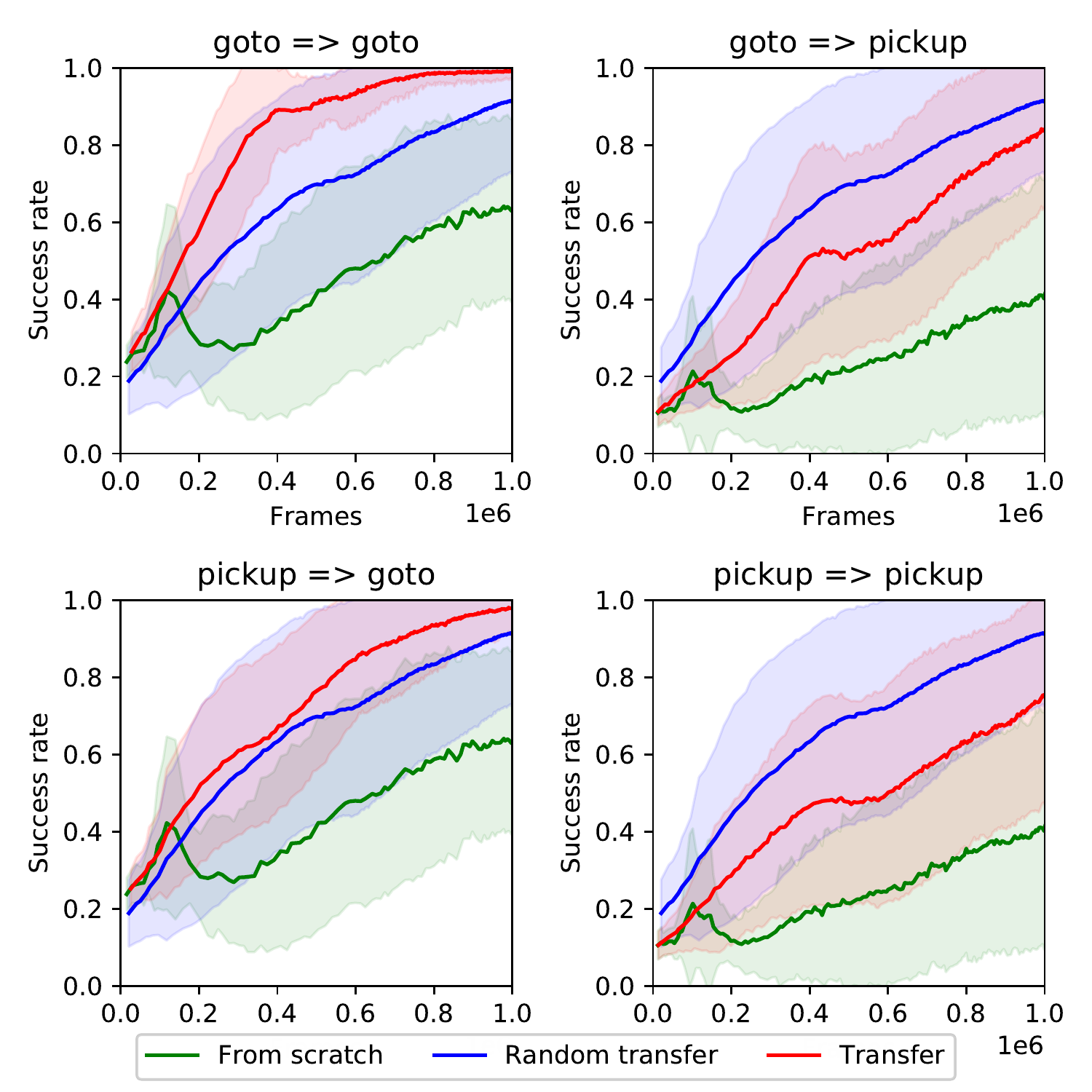}
        \caption{Comparison of how well different base control policies adapt to new tasks, based on whether the verb in the instruction is the same or different.}
        \label{fig:transfer_verbs}
    \end{figure}
    
    \begin{figure}[p]
        \centering
        \includegraphics[scale=0.45]{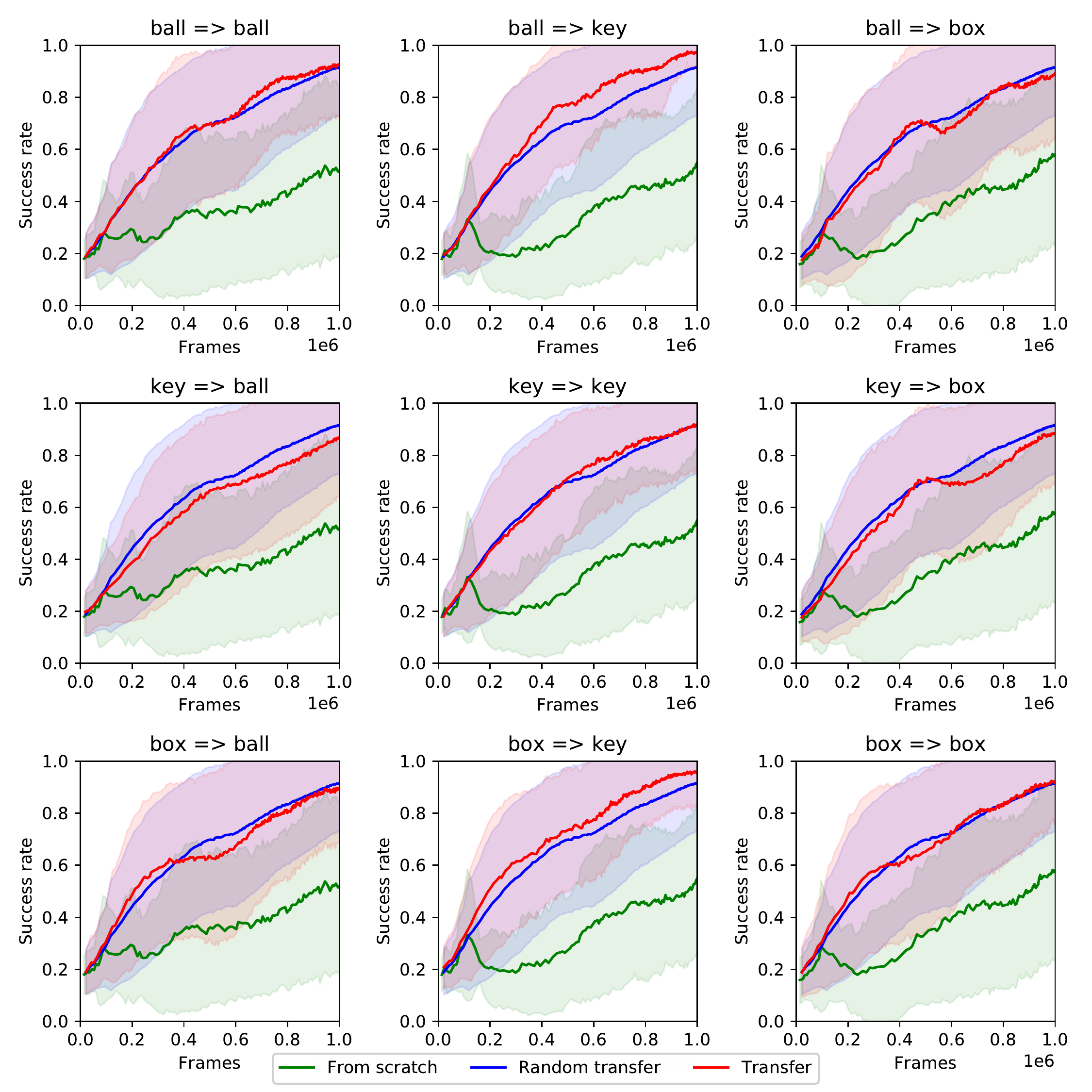}
        \caption{Comparison of how well different base policies adapt to new tasks, based on whether the object in the instruction is the same or different.}
        \label{fig:transfer_objects}
    \end{figure}
    
    \begin{figure}[h]
        \centering
        \includegraphics[scale=0.45]{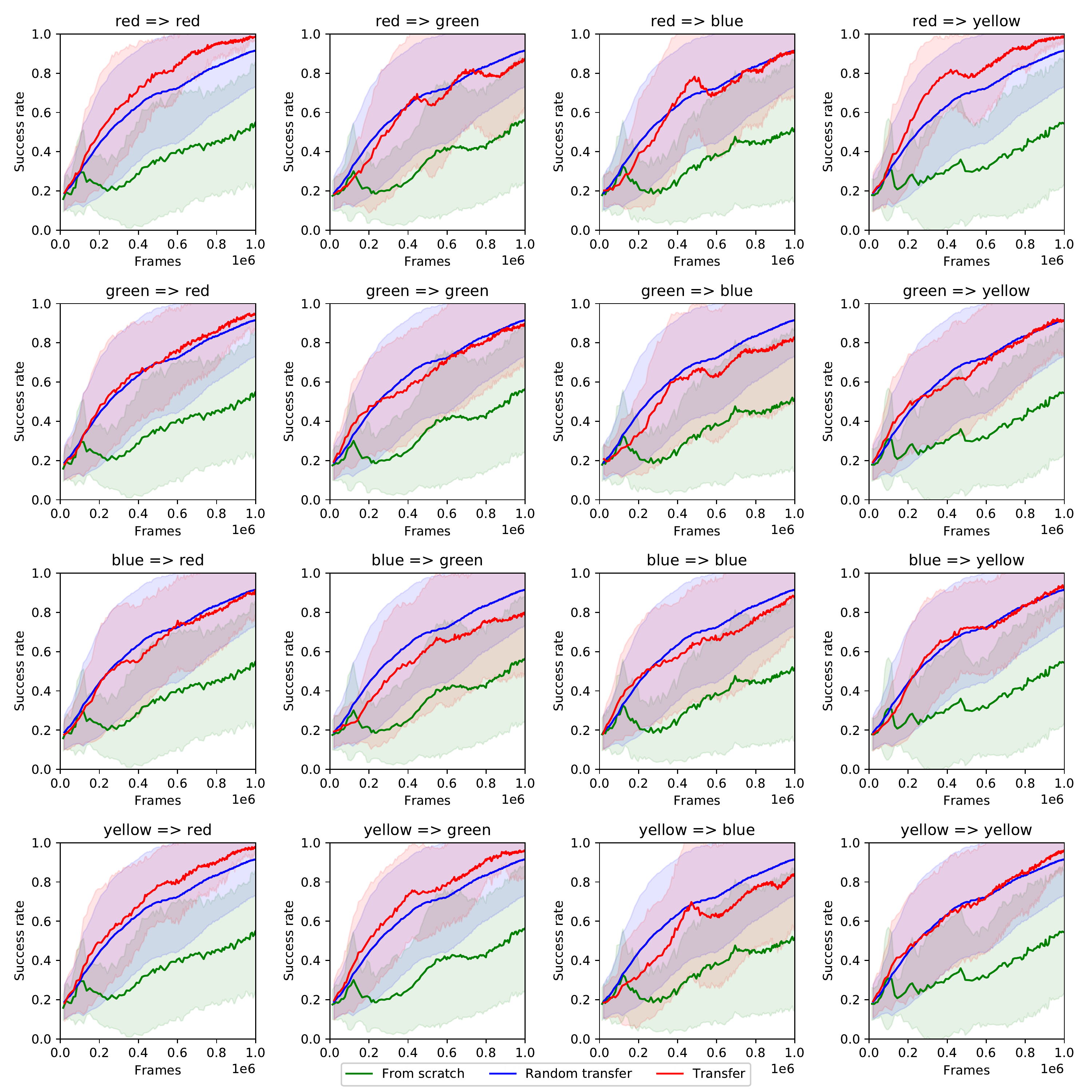}
        \caption{Comparison of how well different base policies adapt to new tasks, based on whether the object color in the instruction is the same or different.}
        \label{fig:transfer_colors}
            
    \end{figure}

    The results of this experiment are summarized in Figure~\ref{fig:transfer_verbs}, \ref{fig:transfer_objects} and \ref{fig:transfer_colors}. Each of these plots shows results averaged over 636 task adaptations. The green line represents performance when training a policy from scratch, while the blue line shows transfer performance averaged over all performed  transfer experiments.
    
    We see some clear patterns. The verb seems to be the most important part of the task instruction. For example, when confronted with a new task which has a \textit{goto} verb, base control policies which are also trained on a \textit{goto} instruction seem to transfer best. This is an expected result, as the verb-part of the instruction, also determines the required set of primitive actions to solve the task.

    \subsection{Transfer model}
    As experimentally demonstrated in the previous experiment, various parts of task instructions have a different impact on the task adaptation performance.

    In this second experiment we trained different amounts $k$ of randomly sampled base control policies. While training can be done using any \gls{rl} algorithm, we used DQN \citep{mnih2015dqn} in our experiments. Training a base control policy is done using at least 1 million steps, and ends when the policy achieves a success rate of at least 95\%, measured on the previous 100 iterations. The full set of used training hyperparameters is described in Appendix~\ref{appendix:hyperparameters}.

    After developing $k$ different base control policies, we sampled $p$ task adaptations for each base control policy. The results gathered from these task adaptations were used to train the transfer model.
    
    In table~\ref{table:transfer_perf}, we show performance of our model when using various numbers of base control policies ($k$), and different numbers of task adaptation samples ($p$). We measure model accuracy over a holdout-set consisting of all possible expressible task-adaptations not seen during sampling.

    \begin{table}[H]
        \centering
        \begin{tabular}{rllllll}
            \toprule
             & \textbf{p=8} & \textbf{p=10} & \textbf{p=12} & \textbf{p=14} & \textbf{p=18} & \textbf{p=20} \\
             \midrule           
             \textbf{k=8} &  0.61 $\pm$0.03 &  0.62 $\pm$0.03 &  0.61 $\pm$0.05 &  0.64 $\pm$0.05 &  0.65 $\pm$0.02 &  0.66 $\pm$0.03 \\
             \textbf{k=10} &  0.62 $\pm$0.03 &  0.62 $\pm$0.05 &  0.64 $\pm$0.06 &  0.62 $\pm$0.04 &  0.66 $\pm$0.03 &  0.67 $\pm$0.02 \\
             \textbf{k=12} &  0.67 $\pm$0.02 &  0.67 $\pm$0.01 &  0.66 $\pm$0.02 &  0.67 $\pm$0.02 &  0.68 $\pm$0.02 &  0.66 $\pm$0.04 \\
             \textbf{k=14} &  0.64 $\pm$0.04 &  0.66 $\pm$0.02 &  0.67 $\pm$0.03 &  0.69 $\pm$0.01 &  0.69 $\pm$0.03 &  0.68 $\pm$0.01 \\
             \textbf{k=18} &  0.67 $\pm$0.03 &  0.68 $\pm$0.02 &  0.68 $\pm$0.03 &  0.71 $\pm$0.01 &  0.70 $\pm$0.02 &  0.71 $\pm$0.02 \\
             \textbf{k=20} &  0.69 $\pm$0.01 &  0.68 $\pm$0.05 &  0.70 $\pm$0.02 &  0.69 $\pm$0.04 &  0.71 $\pm$0.03 &  0.71 $\pm$0.03 \\
            
            \bottomrule
            \\
            \end{tabular}
            
        \caption{Accuracy of the binary task adaptation classifier model. The different rows represent the various amount of base control policies used during training, the columns represent the amount of task adaptations sampled for each base control policy. Results are averaged over 5 runs.}
        \label{table:transfer_perf}
    \end{table}

    Our preliminary results show that even with a limited number of base control policies, and sampled task adaptations, a transfer model can be developed. There is still room for improvement regarding the accuracy of the model, however the stochastic nature of \gls{rl} makes task transfer inherently noisy.

    However the increased sample efficiency, due to efficient task-adaptation provided by our method, is a quintessential building block, in a lifelong learning setting \citep{silver2013lifelong}.

\section{Discussion and future work}
\label{sec:discussion}
In this paper, we presented a method capable of predicting, given a set of base control policies, which of these base control policies will adapt the fastest to a new previously unseen task. In order to make assessments about task adaptation, our method uses a for this task specifically trained language embedding on the task instructions.

Our preliminary results show that a binary classification approach can make assessments about task-adaptation by utilizing semantic meaning of task instructions formatted in natural language. When confronted with an expanding set of tasks in a lifelong-learning setting, our method has the potential to vastly improve sample efficiency.

However, our method still relies on a set of randomly selected base control policies, and task transfer samples. Future research could optimize our method by introducing an iterative sampling method based on a more advanced selection criterion such as \textit{instruction diversity}. Another interesting extension to our method includes the usage of an open vocabulary.

\bibliography{language_transfer}

\appendix
\section{Training hyperparameters}
\label{appendix:hyperparameters}

The network architecture used corresponds to the proposed architecture in \citet{mnih2015dqn}:

\textbf{Network architecture:}

\begin{verbatim}
Relu:Linear(in_features: 512, out_features: 7)
Relu:Linear(out_features: 512)
Relu:Conv(in_channels:64, out_channels:64, kernel_size=3, stride=1)
Relu:Conv(in_channels:32, out_channels:64, kernel_size=4, stride=2)
Relu:Conv(in_channels:3, out_channels:32, kernel_size=8, stride=4)
\end{verbatim}

\textbf{Training hyperparameters:}

\begin{itemize}
    \item Experience replay size: 100.000
    \item Discount factor $\gamma$: 0.99
    \item Adam learning rate: 0.0000625
    \item Adam $\epsilon$: 0.00015
    \item Target-network update steps: 8.000
    \item Random exploration steps: 10.000
    \item Exploration decay-steps: 1.000.000
    \item Minimum exploration $\epsilon$: 0.01
\end{itemize}

During base policy training, the policy is trained until convergence. For the task adaptation sampling 100.000 training steps are used for each task adaptation sample.

\section{Transfer model architecture and hyperparameters}
\label{appendix:transfer_model}

\textbf{Network architecture:}
\begin{verbatim}
Sigmoid:Linear(in_features: 24, out_features: 1)
Relu:Linear(in_features: 24, out_features: 24)
Relu:Linear(in_features: 9, out_features: 24)
Dropout(p=0.2)
Embedding(num_embeddings: 10, embedding_dim: 1)
\end{verbatim}

\textbf{Network hyperparameters:}

\begin{itemize}
    \item Training steps: 1.000.000
    \item Adam learning rate: 0.001
\end{itemize}

\end{document}